 \documentclass[pmlr,twocolumn,10pt]{jmlr} 

\usepackage{booktabs}
\usepackage{siunitx}

\usepackage{caption} 

\usepackage{graphicx}
\usepackage{makecell}
\usepackage{multirow}
\usepackage{colortbl}
\usepackage{xcolor}
\usepackage{listings}
\usepackage{tcolorbox}
\usepackage[switch]{lineno}

\newcommand{\equal}[1]{{\hypersetup{linkcolor=black}\thanks{#1}}}

\jmlrvolume{259}
\jmlryear{2024}
\jmlrsubmitted{LEAVE UNSET}
\jmlrpublished{LEAVE UNSET}
\jmlrworkshop{Machine Learning for Health (ML4H) 2024} 

 \title[RadFlag: Black-Box Hallucination Detection for Medical VLMs]{RadFlag: A Black-Box Hallucination Detection Method for Medical Vision Language Models}

\author{%
\Name{Serena Zhang}\equal{These authors contributed equally.} \Email{serena2z@stanford.edu}\\
\addr Department of Computer Science, Stanford University \\ 
\addr Department of Biomedical Informatics, Harvard Medical School 
\AND
\Name{Sraavya Sambara}\footnotemark[1] 
\Email{sraavya\_sambara@college.harvard.edu} \\
\addr Department of Biomedical Informatics, Harvard Medical School
\\
\Name{Oishi Banerjee} \Email{oishi\_banerjee@g.harvard.edu}\\
\addr Department of Biomedical Informatics, Harvard Medical School \\
\Name{Julian N. Acosta} \Email{julian\_acosta@hms.harvard.edu}\\
\addr Department of Biomedical Informatics, Harvard Medical School \\
\Name{L. John Fahrner} \Email{fahrner@gmail.com}\\
\addr Midwest Radiology \\
\Name{Pranav Rajpurkar} \Email{pranav\_rajpurkar@hms.harvard.edu}\\
\addr Department of Biomedical Informatics, Harvard Medical School
}
\begin{document}

\maketitle
\begin{abstract}
Generating accurate radiology reports from medical images is a clinically important but challenging task. While current vision language models show promise, they are prone to generating hallucinations, potentially compromising patient care. We introduce RadFlag, a black-box method to enhance the accuracy of radiology report generation. Our method uses a sampling-based flagging technique to find hallucinatory generations that should be removed. We first sample multiple reports at varying temperatures and then use a large language model to identify claims that are not consistently supported across samples, indicating that the model has low confidence in those claims. Using a calibrated threshold, we flag a fraction of these claims as likely hallucinations, which should undergo extra review or be automatically rejected. Our method achieves high precision when identifying both individual hallucinatory sentences and reports that contain hallucinations. As an easy-to-use, black-box system that only requires access to a model's temperature parameter, RadFlag is compatible with a wide range of radiology report generation models and has the potential to broadly improve the quality of automated radiology reporting.
\end{abstract}


\paragraph*{Data and Code Availability} The primary dataset used in this research is MIMIC-CXR, an open-source chest X-ray report generation dataset publicly available through PhysioNet \citep{johnson2019mimic}. Code will be made publicly available at \url{https://github.com/rajpurkarlab/RadFlag}.

\section{Introduction}
\label{sec:intro}
Generating accurate radiology reports from medical images is a challenging task that requires a deep understanding of medical imaging and the ability to precisely interpret and communicate complex findings. Medical Vision Language Models (VLMs) that automatically generate radiology reports from chest X-ray images have the potential to mitigate shortages of radiological expertise and improve clinical efficiency and accuracy \citep{bannur2024maira2groundedradiologyreport,Tanida_2023, chen2024chexagentfoundationmodelchest}. However, such models are prone to hallucinations, where the system generates incorrect claims that are not supported by the input image \citep{liu2024surveyhallucinationlargevisionlanguage, bai2024hallucinationmultimodallargelanguage}. For example, even high-performing radiology report generation models can hallucinate in approximately 40\% of generated sentences (see Table 2 in Appendix \ref{apd:fourth}). Such errors can mislead clinicians, with potentially severe consequences for patient care. 

To reduce the adverse impact of hallucinations, we propose RadFlag, a new, easy-to-use method for detecting hallucinations in AI-generated radiology reports without needing access to the AI model's inner workings. RadFlag is the first black-box hallucination detection method specifically designed for VLMs, and flags both sentences and reports as containing hallucinated findings. We provide two key contributions:

\begin{figure*} 
    \centering 
    \includegraphics[width=\textwidth]{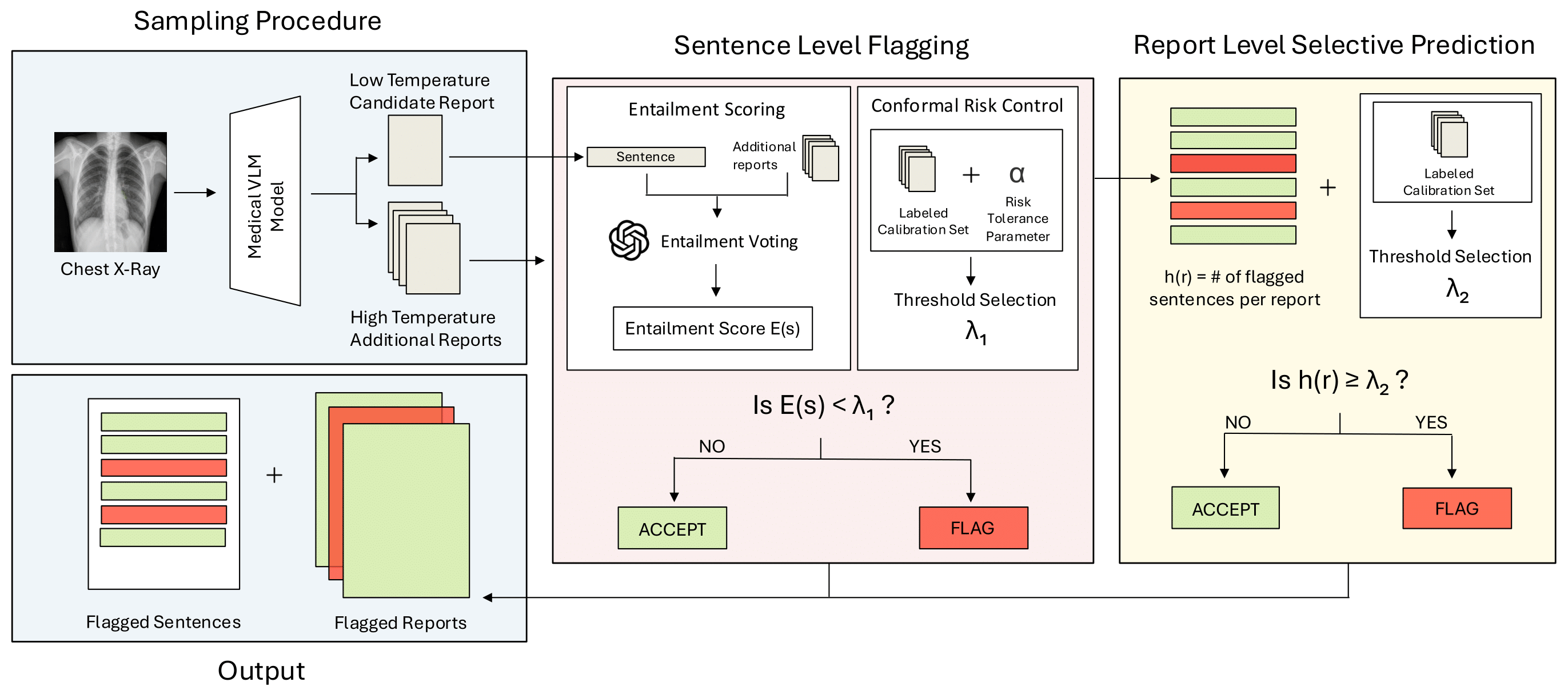} 
    \caption{RadFlag flags hallucinatory text in a candidate report by comparing it against a corpus of additional reports sampled at a higher temperature. GPT-4 is used to count how many of the high-temperature reports support each sentence in the candidate report. We flag sentences with little support, as well as reports that contain multiple flagged sentences.} 
    \label{aba:fig1} 
\end{figure*} 

\begin{itemize}
\item We develop a novel entailment scoring method specifically designed for radiology, which accurately identifies medical findings that a VLM has low confidence in. We combine this scoring method with conformal risk control thresholding to detect hallucinated sentences with high precision. 
\item  We further aggregate the sentence-level flagging results to identify reports with a high rate of hallucinations. This analysis enables selective prediction at the report level, making it possible to abstain from generating reports when there is a high chance of hallucinating. 
\end{itemize}
Our empirical results show that RadFlag can accurately flag 28\% of hallucinatory sentences while maintaining a flagging precision of 73\% on Medversa, a recent high-performing report generation model \cite{zhou2024generalistlearnermultifacetedmedical}. At the report level, our method analyzed 208 reports generated by MedVersa and divided them into two sets: a flagged set with 4.2 hallucinations per report (n = 57) and an accepted set with only 1.9 hallucinations per report (n = 151). This approach can unlock selective prediction, where models refrain from generating reports in cases of heightened uncertainty.

\section{Related Work}

\subsection{Uncertainty Modeling and Hallucination Mitigation}

Techniques for uncertainty modeling in LLMs fall into three categories based on how much access to the model is required: white-box (using internal components such as intermediate activations), gray-box (using output probabilities), and black-box (using only generated outputs) \citep{liu2024uncertaintyestimationquantificationllms, huang2023lookleapexploratorystudy}. Recently, black-box methods measuring self-consistency have emerged as accurate, easy-to-use approaches to assess model output confidence \citep{zhang2024luqlongtextuncertaintyquantification, mündler2024selfcontradictoryhallucinationslargelanguage}. Self-Check GPT, a black-box hallucination detection method, compares multiple sampled responses, measures their consistency using tools like BERTscore and LLM prompting, and labels claims that appear inconsistently across samples as hallucinations. Despite using only model outputs, this black-box method demonstrates superior performance compared to gray-box approaches \citep{manakul-etal-2023-selfcheckgpt}. Discrete semantic entropy offers another black-box approach that clusters multiple generated outputs based on semantic similarity and computes entropy over these clusters to measure uncertainty \citep{farquhar2024detecting}. While these methods have proven effective for text-only models, to the best of our knowledge, RadFlag is the first work to extend this concept to VLMs.

In the domain of radiology report generation, various techniques have been explored to mitigate hallucinations and improve accuracy. Existing methods are generally not black-box and often involve fine-tuning a model with new losses or datasets \citep{banerjee2024directpreferenceoptimizationsuppressing, anonymous2024hallucination, ramesh2022improving}. Additionally, the intersection of uncertainty modeling and hallucination detection remains relatively unexplored for radiology report generation. To the best of our knowledge, only one study has applied uncertainty modeling to remove hallucinations in radiology report generation by using Monte Carlo dropout to estimate uncertainty at both the image and text levels, integrating these estimates into a novel loss function \citep{wang2024trust}. While effective, this method relies on access to model weights. In contrast, our sampling-based method provides a versatile, easy-to-use solution that is model-agnostic and compatible with both proprietary and closed-source models.

\subsection{Selective Prediction for Abstention}

The concept of selective prediction, where models abstain from generating predictions when there is a high chance of error, has also been explored in Natural Language Processing (NLP) \citep{geifman2017selective, xin-etal-2021-art, vazhentsev-etal-2023-hybrid}. Selective prediction relies on uncertainty estimation to decide whether to withhold a response, using the assumption that the model is likely to be incorrect when uncertain. A study by \citet{yadkori2024mitigatingllmhallucinationsconformal} reduces hallucinations in LLM question-answering by combining selective prediction with self-consistency sampling. They evaluate response similarity across samples and use a calibration dataset to set a selective prediction threshold, offering strong theoretical guarantees on error rates. Selective prediction has also been applied to medical imaging. In chest X-ray image classification, selective prediction has shown to improve overall accuracy when models abstain when uncertain about chest X-ray findings \citep{ghesu2021quantifying}. We are the first to develop domain-specific uncertainty estimation techniques for selective prediction in radiology report generation.

\section{Methods}

The RadFlag method is designed to flag hallucinatory content in generated radiology reports as depicted in Figure \ref{aba:fig1}. The method consists of the following key steps:

\begin{enumerate}
    \item Given an input chest X-ray image and a report generation model, we generate one candidate report at a low temperature setting. We also generate a corpus of reports sampled at a higher temperature setting.  
    \item We tokenize the candidate report into sentences. For each sentence, we compute an entailment score using GPT-4 \citep{achiam2023gpt}, where this score represents the number of sampled reports in the high-temperature corpus that support this particular sentence. 
    \item The computed entailment scores for each sentence are compared against a sentence-level threshold, which is chosen using a calibration dataset. Sentences with scores below this threshold are flagged as likely hallucinations.
    \item We tally the number of flagged sentences in each candidate report. The total number of flagged sentences in each candidate report is compared against a report-level threshold, also determined using the calibration dataset. If the count exceeds this report-level threshold, the candidate report is flagged as being especially likely to contain hallucinations.
\end{enumerate}

\subsection{Radiology-Specific Entailment Function}
To determine whether a sentence \( s \) is entailed by a report \( r \), we define a few-shot entailment function tailored for GPT-4 using radiology-specific knowledge. The detailed entailment prompt can be found in Appendix \ref{apd:first}.
 We categorize entailment into three key groups:

\begin{itemize}
    \item \textbf{Completely Entailed (CE):} This category includes sentences that are fully confirmed by the report. For example, the sentence ``atelectasis in the left lung" would be confirmed by a report mentioning ``bilateral atelectasis." Notably, negative findings, like ``the lungs are clear" or ``there are no bony abnormalities," are also classified as CE even when not explicitly confirmed by the report, as it is common clinical practice to leave negative findings implied rather than exhaustively list them.

    \item \textbf{Partially Entailed (PE):} This category captures sentences that are mostly confirmed by the report but have slightly different details, such as severity, size, or anatomical location. For example, the sentence ``there is mild cardiomegaly" would be partially entailed by a report that mentions ``moderate cardiomegaly." Additionally, this category is used when entailment is unclear due to references to prior exams, such as when a sentence states ``heart size is normal" and a report states ``heart size is unchanged."

    \item \textbf{Not Entailed (NE):} This category includes findings that are either directly contradicted by the report or, in the case of positive findings, omitted entirely. For example, if the sentence claims ``there is no pneumothorax" but the report states ``there is a pneumothorax," the sentence is NE. Furthermore, if the sentence claims ``there is consolidation" but the report does not mention consolidation, the sentence is NE, as reports typically list all abnormalities that are present.
\end{itemize}

\subsection{Entailment Scoring for Candidate Sentences using High-Temperature Reports}

We use an entailment score to represent how much support each sentence in the candidate report, which is sampled at a low temperature, has from the corpus of additional reports sampled at a fixed higher temperature. We define this score as $E(s, \{r_1, \dots, r_n\})$ where \( s \) is a sentence from the candidate report, and \( r_i \) is the $i$th report in the corpus of \( n \) high-temperature samples. To compute this score for a sentence $s$, we use the entailment function described above to compare $s$ against every high-temperature report, and we then sum the number of reports categorized as ``Completely Entailed" or ``Partially Entailed." The final score can range from 0, indicating that no high-temperature report samples support the sentence, to $n$, indicating that all sampled reports support the sentence.

In this step, we choose to count ``Partially Entailed" sentences as entailed because models often make small mistakes at higher temperatures despite being correct at lower temperatures. For example, we observe a model correctly noting that an endotracheal tube is 2.5 inches from the carina in a candidate report but claiming that it is 3.5 inches away in a high-temperature sample. To maintain precision and avoid incorrectly flagging factual sentences as hallucinations, we combine ``partial entailment" with ``complete entailment" when comparing sentences against high-temperature samples; we only categorize a sentence as ``not entailed" when there is a substantive mismatch in content. The final score can range from 0, indicating that no high-temperature report samples support the sentence, to $n$, indicating that all samples support the sentence.

\subsection{Sentence-Level Flagging Method}

\begin{equation}
f_{\text{sent}}(s; r_1, \dots r_n; \lambda_1) =
\begin{cases} 
1 & \text{if } E(s, \{r_1, \dots r_n\}) < \lambda_1, \\
0 & \text{if } E(s, \{r_1, \dots r_n\}) \geq \lambda_1.
\end{cases}
\end{equation}

When flagging sentences, the entailment score for each sentence is compared against a calibrated threshold \(\lambda_1\). If the entailment score falls below this threshold, the sentence is flagged as a likely hallucination, as it is poorly supported by the high-temperature reports. Otherwise, the sentence is accepted, not flagged.

\subsection{Calibrating Sentence-Level Thresholds Using Conformal Risk Control}

We employ a Conformal Risk Control (CRC) framework to choose \( \lambda_1 \), our sentence-level hallucination threshold, using an automatically labeled calibration dataset. CRC provides rigorous theoretical guarantees on performance by controlling how often factual sentences are incorrectly flagged as possible hallucinations. Informally speaking, it aims to keep the proportion of factual statements that are flagged by our method from exceeding \( \alpha \), where \( \alpha \) is a parameter that users set based on their risk tolerance. Using a low value for \( \alpha \) allows us to achieve high precision when flagging.

To label our calibration set, we define a binary entailment function \( E_{cal}(s, r_{gt}) \). \( s \) is a sentence from the candidate report, which was generated by a VLM at a low temperature, and \( r_{gt} \) is the corresponding expert-written ground-truth report. As above, we apply our GPT-4 entailment function and treat a sentence as entailed if GPT-4 chooses CE and as not entailed if GPT-4 chooses PE or NE, indicating any mismatch in content. If a sentence is entailed, \( E_{cal}(s, r_{gt}) \) = 1; otherwise, \( E_{cal}(s, r_{gt}) \) = 0.

We then define a loss function \( l(s;r_{gt};\lambda) \) to calibrate the sentence-level threshold \( \lambda_1 \). This loss function is designed to penalize the incorrect flagging of non-hallucinated sentences. The loss function is expressed as:
\begin{align}
l(s;r_{gt}; \lambda) = f_{\text{sent}}(s; r_1, \dots, r_n; \lambda_1) \cdot E_{cal}(s, r_{gt})
\end{align}

This loss function equals 1 when a factual sentence entailed by the ground truth is incorrectly flagged, and it equals 0 otherwise. We next follow the CRC framework to choose a threshold \( \lambda_1 \) that will control how often factual sentences are incorrectly flagged. To do so, we set \( \lambda_1 \) as follows:

\begin{align}
\lambda_1 = \inf \left\{ \lambda : \frac{c+1}{c}L_c(\lambda) + \frac{1}{c+1} \leq \alpha \right\}
\end{align}

where for simplicity, $L_c(\lambda)$ represents the average loss on the calibration set and $c$ is the size of the calibration set. For more theoretical details, please see the original paper on CRC \citep{angelopoulos2023conformalriskcontrol}. 

\subsection{Report-Level Selective Prediction}

To flag problematic reports, we start by counting the number of flagged sentences in each candidate report. This count is denoted as \( h(r) \) for candidate report $r$. We compare each report's count against a a report-level threshold \( \lambda_2 \) to decide if the report should be flagged:

\begin{align}
f_{rep}(r, \lambda_2) = \begin{cases} 
1, & \text{if } h(r) \ge \lambda_2 \\
0, & \text{otherwise} 
\end{cases}
\end{align}

Reports with more than $\lambda_2$ hallucinations are flagged as being especially likely to contain hallucinations, while the rest are accepted. The value of $\lambda_2$ can be chosen based on its behavior on the calibration set; we explore values of $\lambda_2$ that flag 5-25\% of candidate reports. 

\subsection{Models and Sampling}

To evaluate the performance of RadFlag, we apply it to Medversa and RaDialog, two of the highest-performing radiology report generation models on ReXrank, a leaderboard developed for evaluating radiology report generation models from medical images \citep{rexrank, zhou2024generalistlearnermultifacetedmedical, pellegrini2023RaDialoglargevisionlanguagemodel}.  Both Medversa and RaDialog leverage LLMs for report generation, allowing for temperature-based sampling as required by our method. Based on model specifications for ideal hyperparameters, we chose temperature = 0.1 for the low-temperature generation. For the high temperature generations, prior literature referenced a range of 0.9-1.0 as a reasonable temperature setting \citep{yadkori2024mitigatingllmhallucinationsconformal, manakul-etal-2023-selfcheckgpt}. After  qualitatively examining generations from both MedVersa and RaDialog at various temperatures, we chose temperature 0.5 and 5.0, respectively, since they resulted in diverse generations which were still stylistically coherent.

\subsection{Datasets}

MIMIC-CXR is a publicly available dataset of chest X-ray images and associated free-text radiology reports \citep{johnson2019mimic}. We use the MIMIC-CXR test set to evaluate our pipeline for models trained on MIMIC-CXR. We curate a subset of 508 image-report pairs (comprising approximately 3,500 sentences), enriched for reports that have few to no references to prior exams (eg. ``the heart is stable") that would make judging entailment especially difficult. We randomly sampled 300 out of 508 reports to create the calibration set. The remaining 208 reports formed our test set. 

\section{Experiments}

\subsection{Choosing $\lambda_1$ as a Threshold for Flagging Sentences}
Applying conformal prediction to determine the desired threshold $\lambda_1$ requires setting $\alpha$, a risk tolerance parameter that controls how often factual sentences are flagged as hallucinations. Previous research has used values as high as 0.1 \citep{yadkori2024mitigatingllmhallucinationsconformal}. For our purposes, we decide on a more conservative alpha value of 0.05 to prioritize precision when flagging hallucinations. $\lambda_1$ is then calculated through the CRC procedure described in Methods. Using this value, we arrive at 6 for $\lambda_1$ for MedVersa and 4 for RaDialog, and we use these values throughout our Results section. We also give results from setting $\alpha$ to 0.02 in Appendix \ref{apd:fourth}.

\subsection{Choosing $\lambda_2$ as a Threshold for Flagging Reports}
$\lambda_2$ represents the number of flagged sentences in a report as detected by RadFlag. Reports with fewer flagged sentences than $\lambda_2$ are kept, while those with $\lambda_2$ errors or more are filtered out. This approach results in varying percentages of reports being filtered from the dataset.  We explore three thresholds for $\lambda_2$. For both models, $\lambda_2$ = 4 provides a conservative threshold, flagging only 5\% of all reports in the test set. $\lambda_2$ = 3 flags around 10\% of all reports in the test set, while $\lambda_2$ = 2 flags 25\%.

\subsection{Metrics}
We assess report quality using four report-level metrics: RadCliQ-v1, RadGraph entity precision, RadGraph entity recall, RadGraph relation precision, and RadGraph relation recall \citep{yu2023evaluating, jain2021radgraph}. Lower RadCliQ-v1 scores indicate better performance, while higher scores indicate better performance on all RadGraph metrics. Of these metrics, RadGraph entity precision serves as the most direct proxy for hallucinations in radiology reports as it measures how many medical entities in the generated report appear in the ground truth. A lower RadGraph entity precision value suggests that hallucinations have introduced unwanted topics into the report.

\subsection{Defining a True Hallucinations Metric}
We introduce a custom metric termed ``true hallucinations per report" to quantify the occurrence of true hallucinations. We use the same entailment function used to label our calibration set, using GPT-4 to assess whether sentences are supported by their corresponding ground-truth reports or not. To check GPT-4's assessments, one radiologist and one senior internal medicine resident, who are listed in the acknowledgements, and one clinician, who is a contributor to the paper, labeled 547 sentences as “entailed” or “not entailed” by a ground-truth report. We then compare GPT-4’s results to the clinician labels and find that GPT-4 achieves 84\% accuracy. A table with a detailed breakdown is shown in Appendix \ref{apd:second}.

\section{Results}
We demonstrate that RadFlag effectively flags hallucinations with high precision in both Medversa and RaDialog. Furthermore, the report-level selective prediction method reliably identifies reports with a high number of hallucinations.

\subsection{RadFlag Flags Hallucinatory Sentences with High Precision}

\begin{figure}[!h]
\centerline{\includegraphics[width=1.0\linewidth]{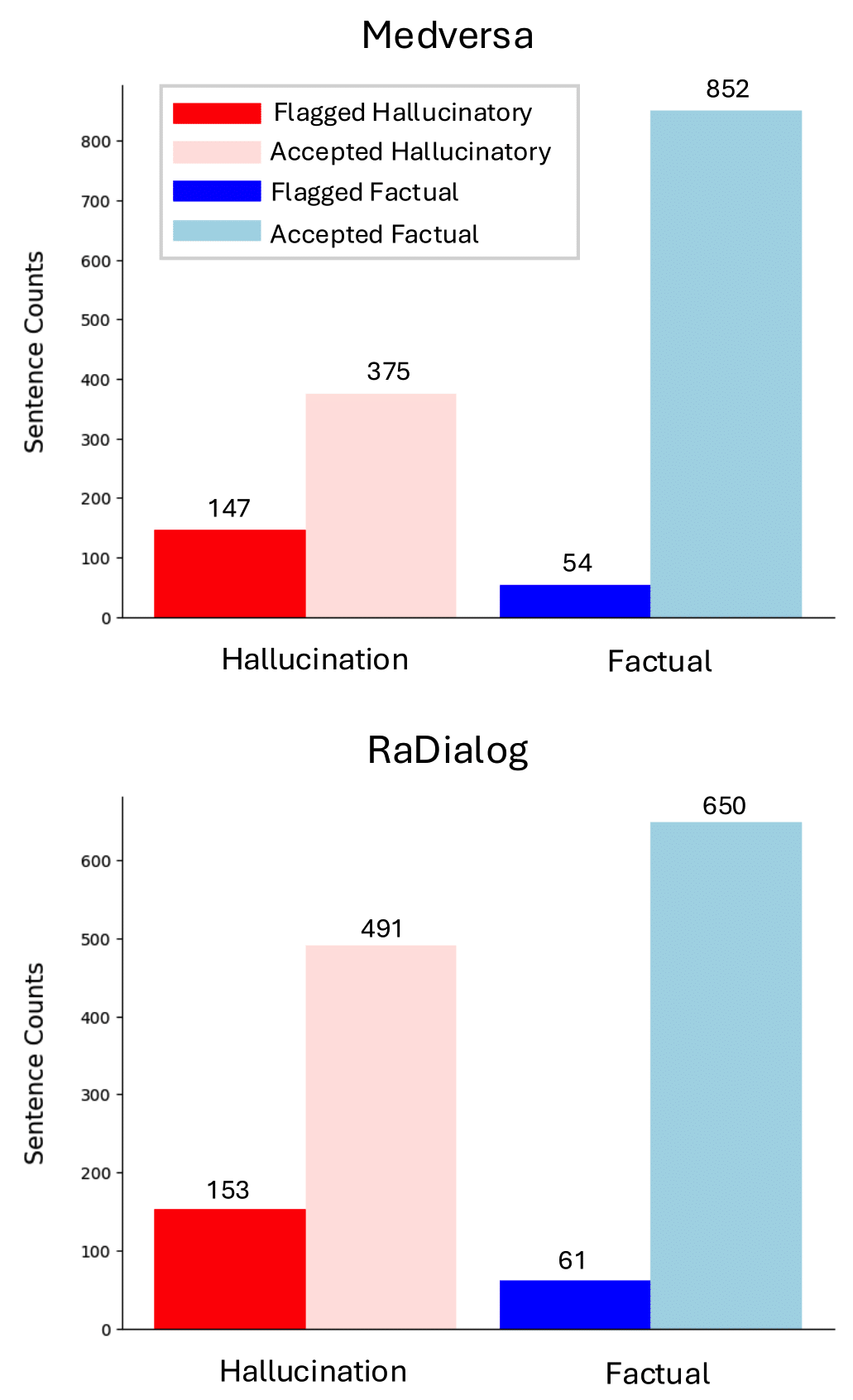}}
\caption{RadFlag flags hallucinatory sentences with high precision for Medversa (top) and RaDialog (bottom).}
\label{aba:fig2}
\end{figure}

Figure \ref{aba:fig2} shows the number of factual and hallucinatory sentences flagged by RadFlag with respect to the remaining factual and hallucinatory sentences. When evaluating Medversa, RadFlag achieves 73\% precision while flagging 28\% of all hallucinations. When evaluating RaDialog, RadFlag achieves 71\% precision while flagging 24\% of all hallucinations.

For a more detailed analysis, we categorized each sentence into one of six medical findings: Lungs, Pleural, Cardiomediastinal, Musculoskeletal, Devices, and Other. These categories were developed with input from radiologists (further details in Appendix \ref{apd:third}).

\begin{figure}[!h]
\centerline{\includegraphics[width=1.0\linewidth]{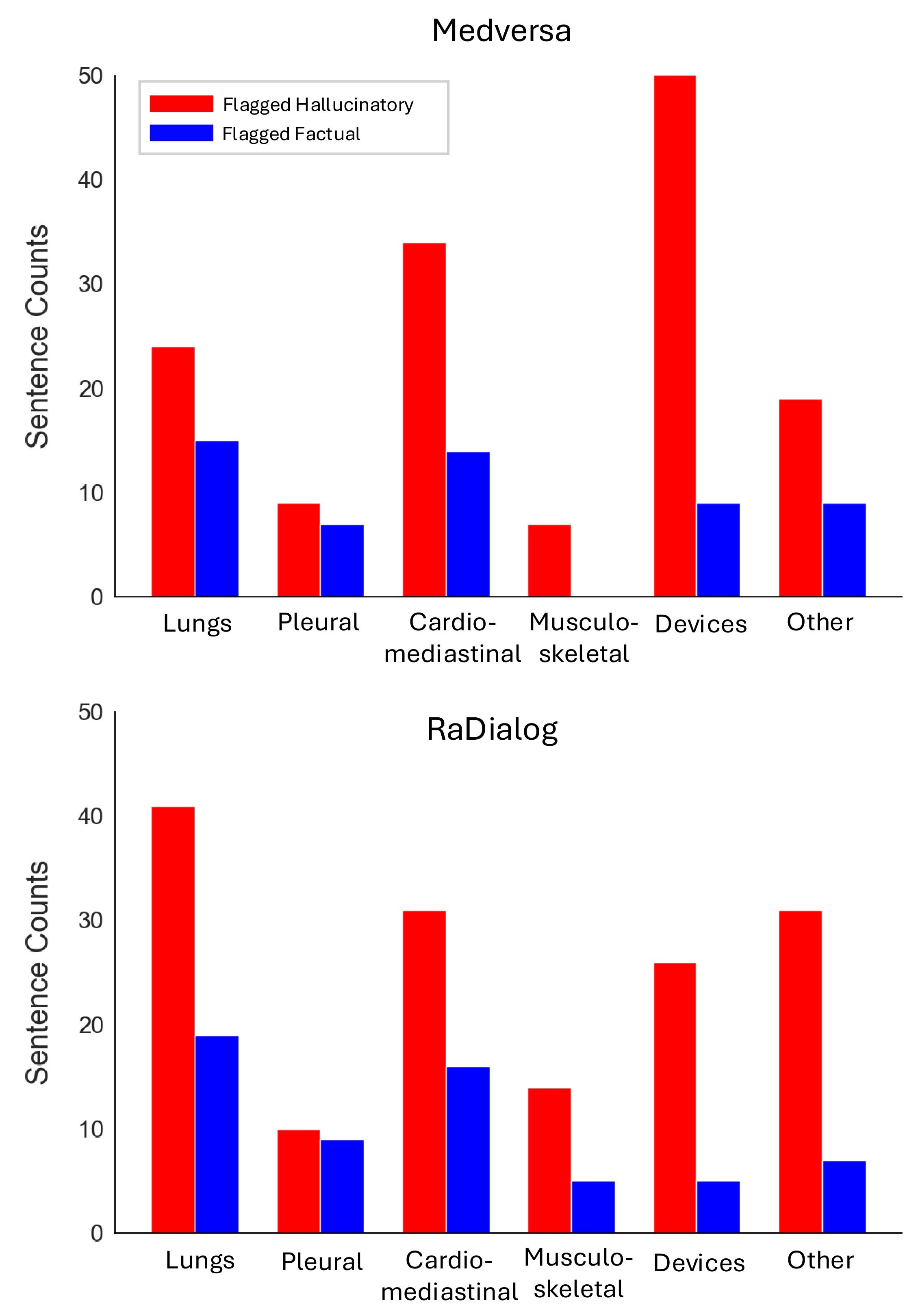}}
\caption{RadFlag consistently flags more hallucinatory sentences than factual sentences, though proportions vary across categories.}
\label{aba:fig3}
\end{figure}

The effectiveness of hallucination detection varies across different findings categories and can be seen in Figure \ref{aba:fig3}. Detailed category counts are provided in Appendix \ref{apd:fifth}. Both models exhibit strong performance in the ``Devices'' and ``Musculoskeletal'' categories, flagging far more hallucinatory than factual statements. However, the sampling method is less effective in the ``Pleural,'' ``Cardiomediastinal,'' and ``Lungs'' categories, where similar numbers of factual and hallucinated sentences are flagged. These differences across categories may reflect underlying variation in how confident and accurate a model is about different topics. This could be because certain features of chest X-ray images are rare in training data and difficult for VLMs to identify, such as device positions or musculoskeletal conditions. 

We evaluate the effect of automatically removing all flagged sentences using RadCliQ-v1 and RadGraph metrics and found the impact on report-level metrics to be relatively minor. This limited effect is likely due to the small number of flagged sentences, as well as the fact that some factual sentences are incorrectly removed in the process. For details on this analysis, please see Appendix \ref{apd:sixth}.

\subsection{RadFlag Flags Lower-Quality, Hallucinatory Reports}

\begin{figure}[!ht]
\centerline{\includegraphics[width=1\linewidth]{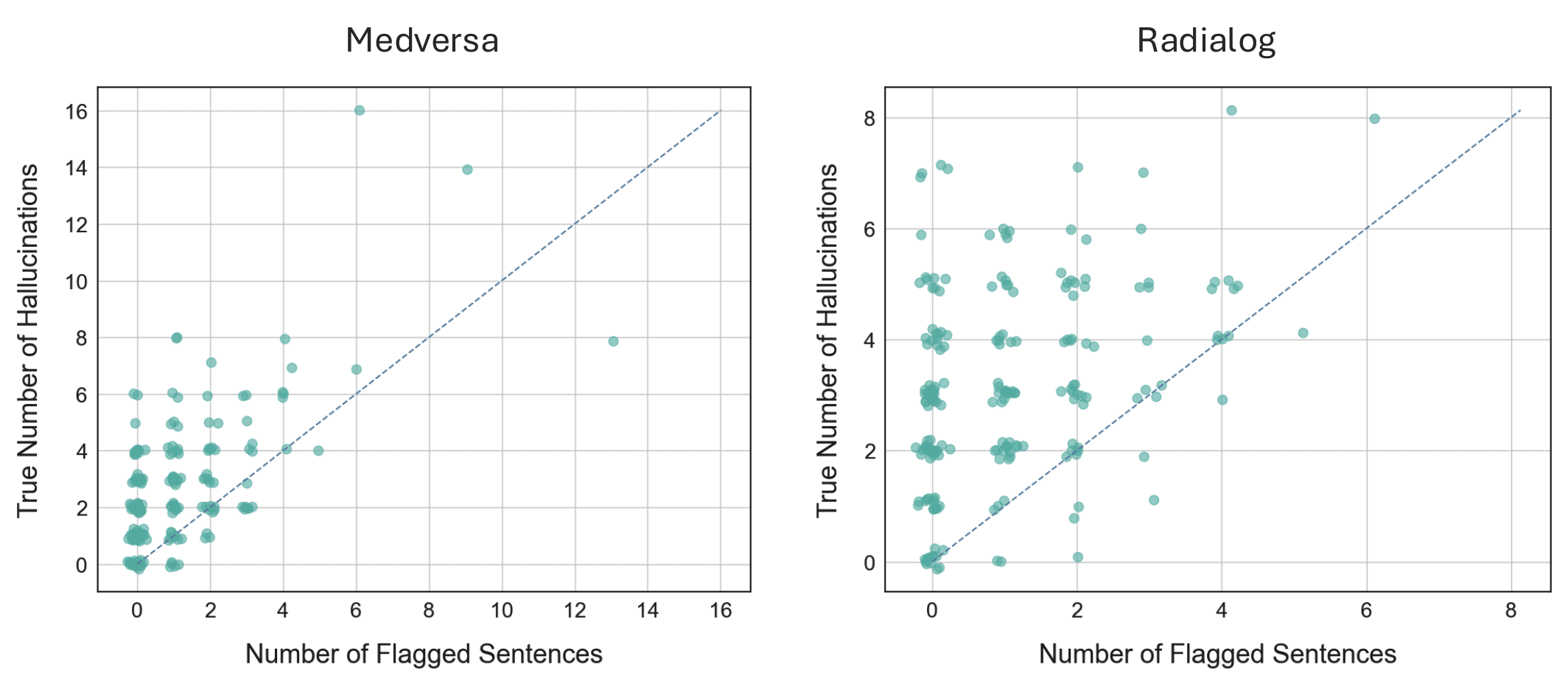}}
\caption{The number of sentences flagged by RadFlag serves as a lower bound for the true number of hallucinations in a report, as the large majority of points fall above the dashed blue line (y=x).}
\label{aba:fig4}
\end{figure}

Next, we aggregate the flagged sentences in each report and use this as a measure to selectively flag reports that are likely to contain multiple hallucinations. Figure \ref{aba:fig4} compares the number of flagged sentences and true hallucinations for reports in our test set, and shows that the majority of the reports fall above the ideal $y = x$ line for both models. For example, Medversa-generated reports with 4 flagged sentences contain at least 4 true hallucinations. Therefore, the count of flagged sentences acts as a reliable lower bound for the number of true hallucinations.

\begin{table*}[htbp]
\centering
\resizebox{\textwidth}{!}{
\begin{tabular}{l l c c c c c c c c} 
\toprule
\textbf{Model} & \textbf{Quality} & \makecell{\textbf{Threshold} \\ \textbf{$\lambda_2$}} & \textbf{n} & \makecell{\textbf{Average True} \\ 
\textbf{Hallucinations} \\ \textbf{per Report}} & \textbf{RadCliQ-v1} & \makecell{\textbf{RadGraph} \\ \textbf{Entity} \\ \textbf{Precision}} & \makecell{\textbf{RadGraph} \\ \textbf{Entity} \\ \textbf{Recall}} & \makecell{\textbf{RadGraph} \\ \textbf{Relation} \\ \textbf{Precision}} & \makecell{\textbf{RadGraph} \\ \textbf{Relation} \\ \textbf{Recall}} \\
\midrule
\multirow{7}{*}{Medversa} & Original   & -- & 208  & 2.5 & 1.085  & 0.379  & 0.275  & 0.206  & 0.130 \\
& Accepted    & 2  & 151  & 1.9 & 1.021  & 0.410  & 0.286  & 0.227  & 0.137 \\
& Flagged & 2  & 57   & 4.2 & 1.290  & 0.279  & 0.241  & 0.137  & 0.110 \\
& Accepted    & 3  & 184  & 2.1 & 1.047  & 0.396  & 0.279  & 0.219  & 0.134 \\
& Flagged & 3  & 24   & 5.4 & 1.376  & 0.249  & 0.244  & 0.103  & 0.101 \\
& Accepted    & 4  & 197  & 2.2 & 1.050  & 0.389  & 0.279  & 0.216  & 0.136 \\
& Flagged & 4  & 11   & 7.8 & 1.630  & 0.192  & 0.195  & 0.020  & 0.034 \\
\midrule
\multirow{7}{*}{RaDialog} & Original   & -- & 208  & 3.1 & 1.222  & 0.308  & 0.241  & 0.140  & 0.103 \\
& Accepted    & 2  & 147  & 2.7 & 1.188  & 0.311  & 0.232  & 0.142  & 0.097 \\
& Flagged & 2  & 61   & 3.6 & 1.213  & 0.299  & 0.262  & 0.135  & 0.120 \\
& Accepted    & 3  & 183  & 2.8 & 1.190  & 0.312  & 0.237  & 0.143  & 0.101 \\
& Flagged & 3  & 25   & 4.0 & 1.233  & 0.275  & 0.271  & 0.119  & 0.118 \\
& Accepted    & 4  & 195  & 2.9 & 1.189  & 0.313  & 0.239  & 0.143  & 0.103 \\
& Flagged & 4  & 13   & 4.6 & 1.290  & 0.269  & 0.099  & 0.107  & 0.336 \\
\bottomrule
\end{tabular}
}
\caption{RadFlag splits reports into flagged and accepted sets, where the flagged reports are consistently lower-quality than accepted reports. Higher scores on True Hallucinations and RadCliQ-v1 indicate worse performance, while higher scores on the RadGraph metrics indicate better performance.}
\label{aba:tab1}
\end{table*}

\begin{figure}[!ht]
\centerline{\includegraphics[width=1.0\linewidth]{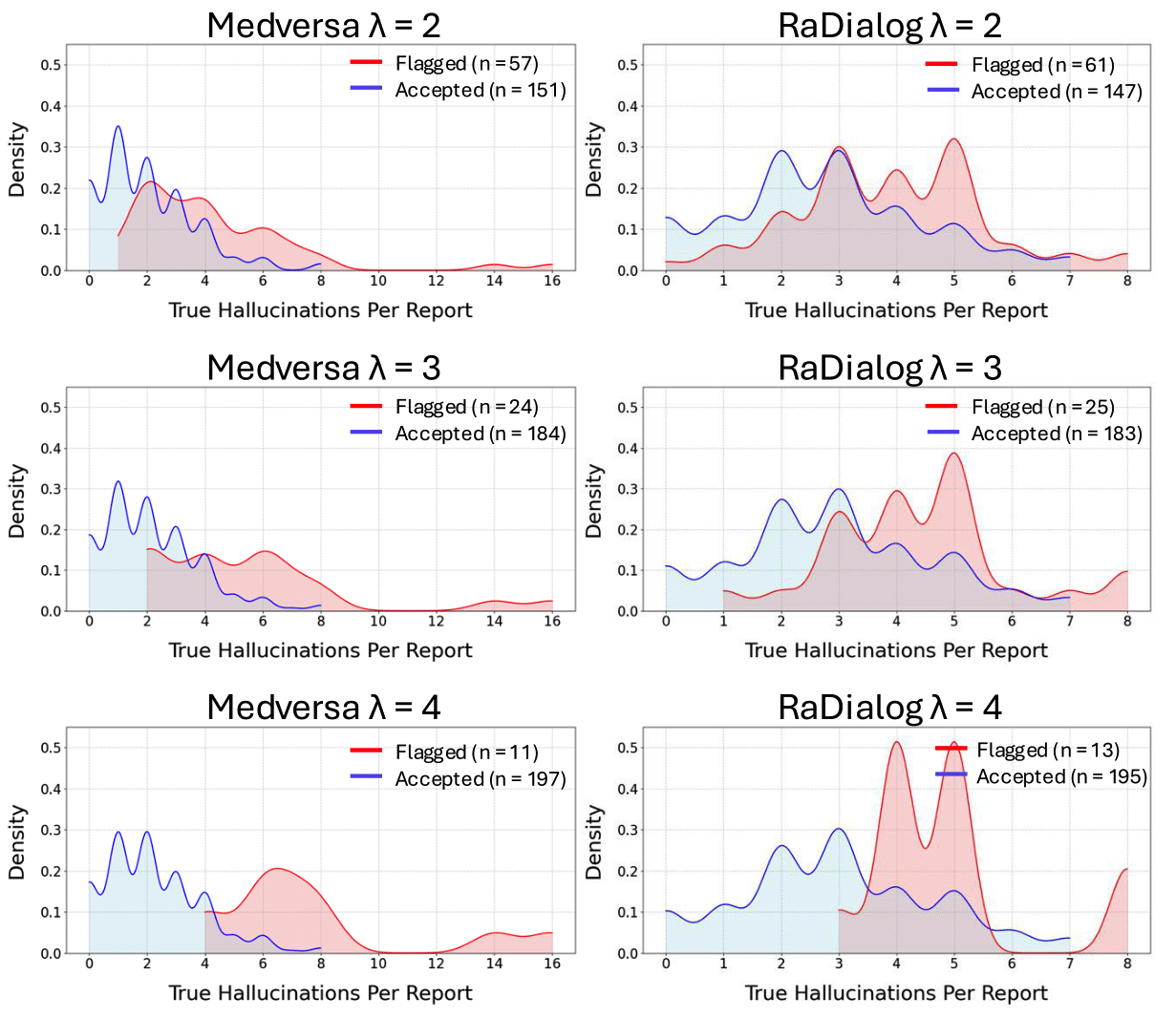}}
\caption{Across models and thresholds, flagged reports tend to have more hallucinations than accepted reports. Nearly all flagged reports contain at least one hallucinatory sentence.}
\label{aba:fig5}
\end{figure}

\begin{figure*}[!ht]
\centerline{\includegraphics[width=\textwidth]{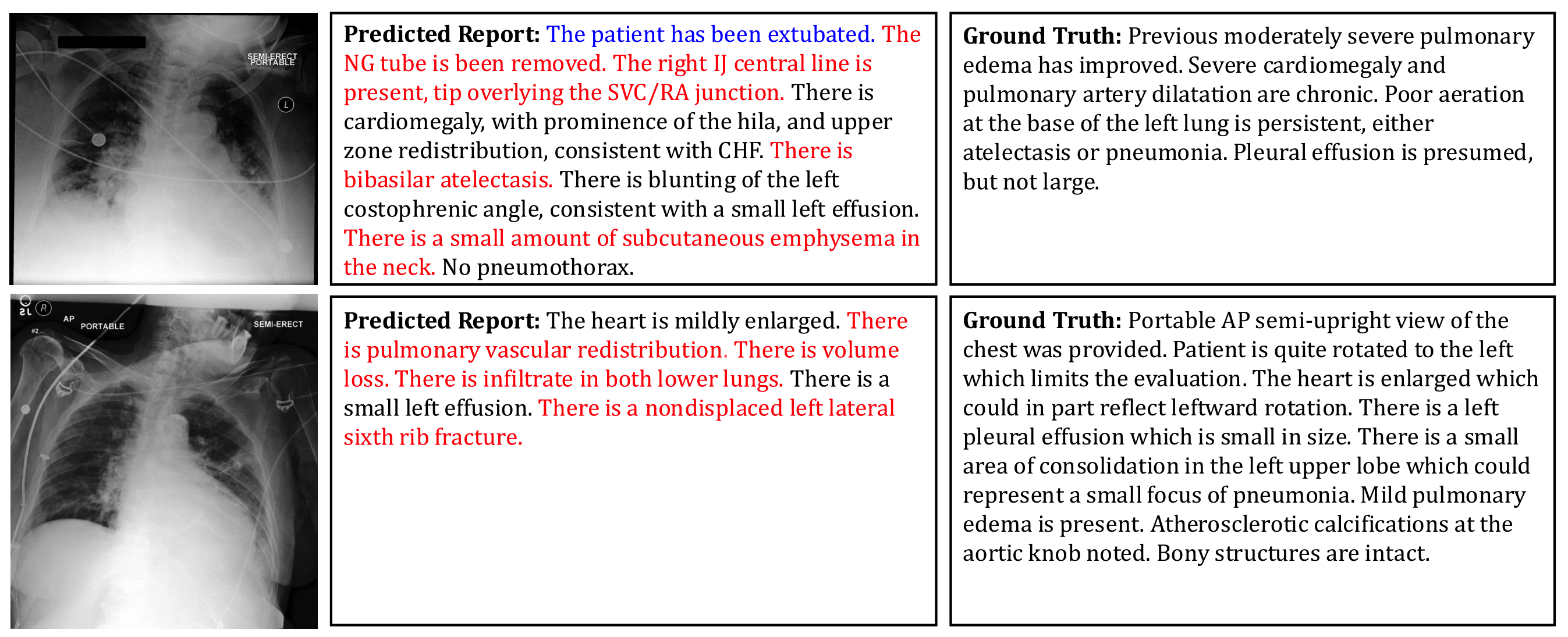}}
\caption{Qualitative examples from Medversa (top row) and RaDialog (bottom row) reports with 4+ hallucinations. Colored text shows radiologist-verified hallucinations: red for those flagged by RadFlag, blue for those missed.}
\label{aba:fig6}
\end{figure*}

Table \ref{aba:tab1} shows that RadFlag successfully flags low-quality reports. For Medversa, there is a considerable disparity in the quality of flagged and accepted reports across all metrics and thresholds. Notably, the 57 reports flagged when $\lambda_2 = 2$  have an average of 4.2 hallucinations per report, while the 151 accepted reports have an average of only 1.9 hallucinations per report. We see a similar trend for RaDialog, though the gaps between flagged and accepted reports are less pronounced. Figure \ref{aba:fig5} shows the distribution of hallucinations in flagged versus accepted reports across both models. According to the true hallucinations metric, RadCliQ-v1, and both RadGraph precision metrics, RaDialog's accepted reports are higher-quality than the flagged reports. In some cases, the flagged reports achieve higher RadGraph recall scores, indicating that these reports contain more hallucinations but also offer better coverage of correct content. 

Our approach also achieves more reliable differentiation between high and low-quality reports, outperforming baseline entropy methods (see Appendix \ref{apd:seventh}). Despite requiring access to per-token probabilities and therefore requiring more technical expertise to implement, we found entropy baselines were not successful in identifying hallucinatory outputs. This is likely because there are many valid ways to express the same radiological idea, thus low probability does not equate to low confidence in the overall claim.

The qualitative examples presented in Figure \ref{aba:fig6} demonstrate the efficacy of the flagging system in identifying  hallucinatory reports from model-generated outputs. This image is included to confirm that the statements are true hallucinations, rather than omissions in the ground truth report.

\section{Discussion}

This study is the first to propose a black-box method for mitigating hallucinations in radiology reports. We introduce RadFlag and evaluate its efficacy when identifying hallucinations produced by two high-performing models. Our approach is capable of flagging individual hallucinated sentences and problematic reports with high precision. 

While our method shows promise in detecting hallucinations, there is room for further improvement. We acknowledge that our reliance on GPT-4 for both entailment scoring and hallucination detection may introduce correlated errors that could inflate our accuracy metrics. While we validated GPT-4's entailment scores with clinicians, future work would benefit from more extensive clinical validation of the hallucination detection process itself. Our findings suggest that category-specific thresholds ($\lambda_1$) for flagging errors could improve performance, given observed variations in our method’s performance across categories. Furthermore, complementary methods could be developed to identify other errors such as omissions, providing a more comprehensive assessment of reports. Finally, an implicit assumption in our current method is that AI models exhibit higher confidence when correct and lower confidence when incorrect. Future methods could test how well this assumption holds for specific models, making it possible to identify models that would most benefit from sampling-based methods.

\section*{Acknowledgments}
We would like to thank Dr. Sara Gomez Villegas and Dr. Jung Oh Lee for their clinical expertise and help in evaluating GPT-4's performance through entailment labeling.

\bibliography{reference}

\newpage

\onecolumn

\appendix

\renewcommand{\thetable}{A\arabic{table}}  
\renewcommand{\thefigure}{A\arabic{figure}}  
\setcounter{table}{0}
\setcounter{figure}{0}

\section{Entailment Prompts}\label{apd:first}
The prompts used for our method can be found in Figures \ref{fig:1} and \ref{fig:2}. All prompting was done using GPT-4. 

\begin{figure}[h]
\centering
\begin{tcolorbox}[colback=white!95!gray, colframe=black!75, width=\textwidth, boxrule=0.5pt, arc=4pt, auto outer arc, boxsep=5pt]
\textbf{Prompt:} 

You are an AI radiology assistant helping process reports from chest X-rays. You are given a finding (F) from a chest X-ray and a list of reports (R1-R10). Your job is to determine if the ground truth report entails the finding, partially entails the finding, or do not entail the finding. Here are the definitions for entailment, partially entail, and do not entail: \\

\textbf{Completely Entailed:} A positive finding is entailed if it is clearly mentioned in the report. It is also entailed if the larger finding (atelectasis in both lungs) in the report includes the finding (atelectasis in the left lung). A negative finding is entailed if it is not mentioned in the report because we can assume that negative findings are not always explicitly stated. Note that negative findings are "normal findings" that include lungs are clear, heart is normal size, contours are normal, heart is not enlarged; even if they're not stated explicitly if it can be inferred that they're normal, then it is entailed. \\

\textbf{Not Entailed:} A positive finding is not entailed when the report doesn't mention the finding at all, or explicitly mentions the complete opposite finding. A negative finding is not entailed when the report mentions the complete opposite finding. \\

\textbf{Partially Entailed:} A positive finding is somewhat entailed if it is close to correct, but not quite– for example, there is a difference in the severity or location of the finding. A device is partially entailed if there are similar (but not the same) kinds of devices in the same position in the reports. Lastly, if there are any priors present in the finding that are absent from the report, it is also somewhat entailed. \\

Here are some examples:

\#1 \textbf{Finding:} "There is no pleural effusion." \textbf{Report:} "There is a pneumothorax. There is no focal consolidation." \textbf{Score:} Completely entailed, because it is a negative finding not mentioned in the report.

\#2 \textbf{Finding:} "The lungs are clear." \textbf{Report:} "No focal consolidation, pleural effusion, or evidence of pneumothorax is seen. There is no overt pulmonary edema." \textbf{Score:} Completely entailed, because "lungs are clear" is a negative finding and the report doesn't mention anywhere that the lungs ARE NOT CLEAR (which only happens if atelectasis, consolidations, pneumonia, or opacities etc. are present). \\

The end of every report will have no additional positive findings. Other negative findings are ALWAYS entailed. \\

Return a json of the format: {"E": [], "P": [], "N": []}, where the list
contains the report number that falls in each category. For example, if a
finding F is completely entailed by reports 1, 2, and 3, and not entailed
by reports 4, 5, and 6, and partially entailed with respect to reports 7,
8, 9, and 10, then the dictionary returned is {"E": [1, 2, 3], "P": [7, 8,
9, 10], "N": [4, 5, 6]}. Note that the length of all E, N, and P MUST ADD
UP to 10.
\end{tcolorbox}
\caption{Prompt used for entailment score function $E(s, \{r_1, \dots, r_n\})$ to compute entailment score.}
\label{fig:1}
\end{figure}

\begin{figure}[!h]
\centering
\begin{tcolorbox}[colback=white!95!gray, colframe=black!75, width=\textwidth, boxrule=0.5pt, arc=4pt, auto outer arc, boxsep=5pt]
\textbf{Prompt:} 
Return a json of the format: \{"status": "entailed"\}, where the status contains completely entailed, partially entailed, or not entailed. For example, if a finding F is not entailed by the report, return \{"status": "not entailed"\}. If there is partial entailment, return \{"status": "partially entailed"\}.
\end{tcolorbox}
\caption{Prompt used for ground truth scoring function $g(s, r)$ to compute True Hallucinations. The entailment scheme used in this prompt mirrors the one in Figure \ref{fig:1}}
\label{fig:2}
\end{figure}

\section{Clinician Entailment Labels}\label{apd:second}
Table \ref{tab:1} verifies that GPT-4 entailment scores align with clinician labels. 

\begin{table}[htbp]
\centering
\begin{tabular}{|l|c|c|}
\hline
 & \textbf{Clinician} & \textbf{Clinician} \\
 & \textbf{Entailed} & \textbf{Not Entailed} \\
\hline
\makecell{\textbf{GPT-4}\\
\textbf{Entailed}} & 279 & 47 \\
\hline
\makecell{\textbf{GPT-4}\\
\textbf{Not Entailed}} & 42 & 179 \\
\hline
\end{tabular}
\caption{Clinician entailment labels align well with our GPT-4 Scoring Function.}
\label{tab:1}
\end{table}

\section{Category Definitions}\label{apd:third}

We categorize each sentence into one of six medical findings to assess how well RadFlag flags hallucinations in each category. These categories were developed in consultation with experienced radiologists to ensure comprehensive coverage of common chest X-ray observations. Below is a list of keywords that comprise each category. If sentences contain keywords from multiple categories, we assign them to the first matching category based on the following priority order: Devices, Cardiomediastinal, Lungs, Musculoskeletal, Pleura, and Other.  We use regular expressions to search for these keywords when going through each sentence:

\begin{itemize}
    \item \textbf{Devices:} PICC, Endotracheal, Nasogastric, Tube, Catheter, Pacemaker, Stent, NG, ET, ETT, Clip, Staple, Coil, ICD, LVAD, RVAD, Tracheostomy, Valve, Plate, Device, Drain, Gastric, Electrode, Lead, Port-a-cath, Cath, Clamp, Defibrillator, Internal Jugular Line, Subclavian Line, Hickman Line, Right Atrial Line, Broviac Line, Pacer, Tip, IJ, NJ, Wires, SVC, Dobbhoff, Intubated, Pump, Port, Extubate, Nasojejunal, Enteric, Impella, Mitraclip, Sternotomy, Suture, Subclavian, IABP, Balloon Pump, Filter, IVC Filter, TAVR, Stent Graft, Pigtail, Large Bore, Chest Tube, Epidural, Stimulator, Fixation, ORIF
    \item \textbf{Cardiovascular:} Hila, Hilar, Hilum, Lymph, Lymphadenopathy, Hypertension, Artery, Arteries, Mediastinal, Aorta, Hernia, Aortic, Mediastinum, Hiatal Hernia, Vascular Plethora, Pneumomediastinum, Cardiomegaly, Pneumopericardium, Pericardial Effusion, Pericardial, Cardio, Cardiac, Heart, Cardium, CHF, Redistribution, Congest, Engorge, Fluid Overload, Vascular, Fluid, Pneumomediastinum, Tension, Air, Shift, Pericardial Effusion
    \item \textbf{Lung:} Inflate, Volume, Bronchovascular, Diaphragm, Emphysema, Hyperinflated, COPD, Low, Hypoinflated, Well Expanded, Overinflated, Flattening, Aerated, Blebs, Cyst, Mass, Nodule, Lesion, Pneumonia, ARDS, Hazy, Haze, Atelectasis, Opacity, Opacit, Consolidation, Edema, Interstitial, Hemidiaphragm, Vessel, Infiltrate, Clear, Fibrosis, Scarring, Scar
    \item \textbf{Musculoskeletal:} Bones, Bony, Demineralized, Compression, Degenerative, Osseous, Fracture, Fractures, Rib, Spine, Scoliosis, Kyphosis, Osteopenia, Osteoporosis, Deformity, Deformities, Listhesis, Shoulder, Dislocation, Clavicle, Scapula, Humeral, Height, Loss, Stable, Unstable, Anterolisthesis, Retrolisthesis, Burst, Extension, Displaced, Distracted, Angulated, Angulation
    \item \textbf{Pleural:} Pneumothorax, Pneumoth, Effusion, Pleural, Thickening, Scarring, Pleural Mass, Costophrenic, Plaque, Blunting, Blunted, Silhouetting
\end{itemize}

\section{Alpha Comparison for Sentence-Level Flagging}\label{apd:fourth}

 Using $\alpha$ values of .02 and .05, we arrive at $\lambda_1$ values of 4 and 6 for MedVersa and 2 and 4 for RaDiolog. Below is a comparison of performance at these two values of $\alpha$.

\begin{table}[htbp]
\centering
\caption{Comparison of Model Performance}
\small
\begin{tabular}{@{}lc>{\centering\arraybackslash}p{1cm}>{\centering\arraybackslash}p{1cm}>{\centering\arraybackslash}p{1cm}>{\centering\arraybackslash}p{1cm}@{}}
\toprule
\textbf{Model} & \textbf{$\alpha$} & 
\textbf{Flag\newline Halluc.} & \textbf{Flag\newline Factual} & \textbf{Accept\newline Factual} & \textbf{Accept\newline Halluc.} \\
\midrule
\multirow{2}{*}{Medversa} & 0.02 & 84 & 19 & 887 & 438 \\
 & 0.05 & 147 & 54 & 852 & 375 \\
\midrule
\multirow{2}{*}{RaDialog} & 0.02 & 84 & 24 & 687 & 560 \\
 & 0.05 & 153 & 61 & 650 & 491 \\
\bottomrule
\end{tabular}
\label{tab:model_comparison}
\end{table}
\newpage

\section{Category-Level Breakdowns}\label{apd:fifth}

 The following charts show the breakdown of flagged hallucinatory, flagged factual, accepted factual, and accepted hallucinatory statements across both models at $\alpha = 0.05$, or $\lambda_1 = 6$ for Medversa and $\lambda_1 = 4$ for RaDialog.

\begin{table}[htbp]
\centering
\caption{Summary of Medversa Findings at $\lambda = 6$ ($\alpha=0.05$)}
\large
\begin{tabular}{@{}lcccc@{}}
\toprule
\textbf{Finding} & \makecell{\textbf{Flag}\\\textbf{Halluc.}} & \makecell{\textbf{Flag}\\\textbf{Factual}} & \makecell{\textbf{Accept}\\\textbf{Factual}} & \makecell{\textbf{Accept}\\\textbf{Halluc.}} \\
\midrule
Lungs & 24  & 15 & 99  & 187 \\
Pleura & 9   & 7  & 63  & 327 \\
Cardiomediastinal & 34  & 14 & 100 & 216 \\
Musculoskeletal & 7   & 0  & 4   & 50  \\
Devices & 54  & 9  & 87  & 33  \\
Other & 19  & 9  & 22  & 39  \\
\bottomrule
\end{tabular}
\label{tab:summary_findings}
\end{table}

\begin{table}[htbp]
\centering
\caption{Summary of RaDialog Findings at $\lambda = 4$ ($\alpha=0.05$)}
\large
\begin{tabular}{@{}lcccc@{}}
\toprule
\textbf{Finding} & \makecell{\textbf{Flag}\\\textbf{Halluc.}} & \makecell{\textbf{Flag}\\\textbf{Factual}} & \makecell{\textbf{Accept}\\\textbf{Factual}} & \makecell{\textbf{Accept}\\\textbf{Halluc.}} \\
\midrule
Lungs & 41  & 19  & 160  & 158 \\
Pleura & 10   & 9  & 60  & 251 \\
Cardiomediastinal & 31   & 16  & 103  & 134 \\
Musculoskeletal & 14   & 5  & 16  & 32  \\
Devices & 26   & 5  & 79  & 39 \\
Other & 31  & 7  & 73  & 36  \\
\bottomrule
\end{tabular}
\label{tab:summary_findings_lambda_4}
\end{table}

\newpage
\section{Effect of Removing Flagged Sentences on Report-level Metrics}\label{apd:sixth}

Table \ref{tab:5} shows that removing flagged sentences from reports does not seem to change overall report-level metric performance. The $*$ indicates adjusted report-level metrics from removing empty reports (reports with all sentences flagged) that inflate RadCliQ-v1 scores.

\begin{table}[htbp]
\centering
\small
\caption{Performance Metrics (Findings Level) for Medversa and RaDialog}
\begin{tabular}{@{}lcccccc@{}}
\toprule
\textbf{Model} & \textbf{Method} & \makecell{\textbf{Avg.}\\\textbf{RadCliQ-v1}} & \makecell{\textbf{Avg. RadGraph}\\\textbf{Entity Precision}} & \makecell{\textbf{Avg. RadGraph}\\\textbf{Entity Recall}} & \makecell{\textbf{Avg. RadGraph}\\\textbf{Relation Precision}} & \makecell{\textbf{Avg. RadGraph}\\\textbf{Relation Recall}} \\
\midrule
\multirow{4}{*}{Medversa} & Original & 1.085 & 0.379 & 0.275 & 0.206 & 0.130 \\
 & $\alpha = 0.02$ & 1.084 & 0.390 & 0.268 & 0.212 & 0.127 \\
 & $\alpha = 0.05$ & 1.156 & 0.395 & 0.256 & 0.215 & 0.120 \\
 & $\alpha = 0.05^*$ & 1.106 & 0.396 & 0.257 & 0.216 & 0.119 \\
\midrule
\multirow{3}{*}{RaDialog} & Original & 1.22 & 0.308 & 0.241 & 0.140 & 0.103 \\
 & $\alpha = 0.02$ & 1.20 & 0.328 & 0.232 & 0.152 & 0.100 \\
 & $\alpha = 0.05$ & 1.21 & 0.344 & 0.221 & 0.163 & 0.094 \\
\bottomrule
\end{tabular}
\label{tab:5}
\end{table}
\newpage
\section{Entropy Baselines}\label{apd:seventh}

To provide a comparison point for our black-box sampling method, we tested two baseline methods used by \cite{manakul-etal-2023-selfcheckgpt} that require access to output probabilities: average negative log probability and average entropy. In an LLM's response, $p_{ij}$ represents the probability of generating the j-th token in the i-th sentence, where J indicates the total token count per sentence. Average negative log probability is calculated as: $$\text{Avg}(-\log p) = -\frac{1}{J} \sum_j \log p_{ij}$$ Entropy is calculated according to the equation $
\mathcal{H}_{ij} = -\sum_{\tilde{w}\in\mathcal{W}} p_{ij}(\tilde{w}) \log p_{ij}(\tilde{w})$.
Average entropy is thus calculated as: $$\text{Avg}(\mathcal{H}) = \frac{1}{J}\sum_j \mathcal{H}_{ij}$$  We ran experiments using a new set of MedVersa generations (temp = 0.1, excluding 11 datapoints where generation failed) on 197 MIMIC-CXR test datapoints, flagging the reports that scored highest on these metrics. Despite requiring access to output probabilities, these baseline methods performed poorly, so hallucinations appear at similar rates in "accepted" and "flagged" sentences.

\begin{table}[htbp]
\centering
\small
\caption{Comparison of different ranking methods and their hallucination detection performance.}
\begin{tabular}{@{}lcccc@{}}
\toprule
\textbf{Ranking Method} & \textbf{Label} & \textbf{n} & \makecell{\textbf{Mean True}\\\textbf{Hallucinations}} & \makecell{\textbf{Median True}\\\textbf{Hallucinations}} \\
\midrule
\multirow{6}{*}{Average Entropy} & Accepted & 187 & 2.63 & 2.0 \\
 & Flagged & 10 & 3.50 & 2.5 \\
 & Accepted & 174 & 2.59 & 2.0 \\
 & Flagged & 23 & 3.30 & 3.0 \\
 & Accepted & 143 & 2.46 & 2.0 \\
 & Flagged & 54 & 3.20 & 3.0 \\
\midrule
\multirow{6}{*}{Average Negative Log Probability} & Accepted & 187 & 2.70 & 2.0 \\
 & Flagged & 10 & 2.10 & 1.50 \\
 & Accepted & 174 & 2.59 & 2.0 \\
 & Flagged & 23 & 3.30 & 3.0 \\
 & Accepted & 143 & 2.46 & 2.0 \\
 & Flagged & 54 & 3.22 & 3.0 \\
\bottomrule
\end{tabular}
\label{tab:ranking_comparison}
\end{table}

\end{document}